\documentclass[11pt,fullpage, letterpaper,twoside]{article}
\usepackage{helvet}
\usepackage{epsfig} 
\usepackage{cite}

\usepackage{algorithm}
\usepackage{algorithmic}
\usepackage{amsmath}
\usepackage{amsthm}
\usepackage{amssymb}
\usepackage{graphicx}
\usepackage{color}
\usepackage{times}
\usepackage{epsfig}
\usepackage{graphicx}
\usepackage{amsmath}
\usepackage{amssymb}
\usepackage{array,multirow}
\usepackage{caption}
\usepackage{subcaption}
\usepackage{dsfont}

\usepackage{multirow}
\usepackage{booktabs}
\usepackage{hhline}
\usepackage{multirow}
\usepackage{url}
\usepackage[margin=1in]{geometry}

\usepackage{CJK}
\usepackage{indentfirst}
\usepackage[colorlinks,linkcolor=red,anchorcolor=blue,citecolor=green]{hyperref}
\usepackage{fancyhdr}
\begin{document}

\title{FLAT: Few-Shot Learning via Autoencoding Transformation Regularizers
}


\author{$^1$Haohang Xu \\
        $^1$Hongkai Xiong, {Senior~Member,~IEEE} \\
        $^2$Guojun Qi, {Senior~Member,~IEEE}    \\
        \textit{$^1$Shanghai Jiao Tong University, Shanghai, China} \\
\textit{$^2$Futurewei Technologies, USA} \\
$^1$Email: \{xuhaohang,xionghongkai\}@sjtu.edu, $^2$Email: guojun.qi@futurewei.com
}



\date{}

\maketitle

\begin{abstract}

One of the most significant challenges facing a few-shot learning task is the generalizability of the (meta-)model from the base to the novel categories.  Most of existing few-shot learning models attempt to address this challenge by either learning the meta-knowledge from multiple simulated tasks on the base categories, or resorting to data augmentation by applying various transformations to training examples. However, the supervised nature of model training in these approaches limits their ability of exploring variations across different categories, thus restricting their cross-category generalizability in modeling novel concepts. To this end, we present a novel regularization mechanism by learning the change of feature representations induced by a distribution of transformations without using the labels of data examples. We expect this regularizer could expand the semantic space of base categories to cover that of novel categories through the transformation of feature representations. It could minimize the risk of overfitting into base categories by inspecting the transformation-augmented variations at the encoded feature level. This results in the proposed FLAT (Few-shot Learning via Autoencoding Transformations) approach by autoencoding the applied transformations. The experiment results show the superior performances to the current state-of-the-art methods in literature.

\end{abstract}


\maketitle

\section{Introduction}\label{sec:intro}

Few-Shot Learning (FSL) \cite{fei2006one,miller2000learning} seeks to build a model to predict on novel categories with few-shot labeled examples from a set of training examples of base categories. Thus, one of fundamental challenges of the FSL problem is to learn a prediction model that can well generalize from the base to the novel categories even with very few examples.

Most of existing FSL methods can be revisited as addressing such a {\em category generalization} problem in different ways. For example, the meta-learning based algorithms such as the LSTM-based meta-learning \cite{Ravi2017}, Model-Agnostic Meta-Learning (MAML) \cite{finn2017model}, and Task-Agnostic Meta-Learning (TAML) \cite{jamal2018task} attempt to train a meta-model by simulating multiple episodes of few-shot learning tasks from base categories. The knowledge about how to adapt to a new concept is learned into the meta-model, so that it can be applied to predict on novel categories with few examples.

While it is a sound idea to learn such a meta-model for generalizing to novel categories, the simpler weight imprinting \cite{qi2018low} and weight generator \cite{gidaris2018dynamic,qiao2018few} approaches can reach the comparable (if not better) performances. weight imprinting directly imprints the prediction weights of novel categories with their mean representations on top of the neural networks pre-trained with labeled examples of base categories. Alternatively, weight generator predict the weights as a function of the feature representations using their means \cite{qiao2018few} or an attention mechanism \cite{gidaris2018dynamic}.  However, an explicit mechanism is lacking to deliver more competitive performances on novel categories by generalizing from base categories in these simple yet effective methods.

To this end, we propose a novel category generalization method via autoencoding transformations. It is inspired by training a generalizable model via a rich family of transformations on images of base categories. However, a naive application of these transformations through data augmentation does not work in this FSL scenario. First, in the naive data augmentation, the model is still pre-trained with the labels of base categories even if their training examples are augmented with various transformations. In the fine-tuning stage, due to the extremely rare examples ($K$-shot) of novel categories, the transformation-augmented examples would not help a lot in delivering better generalization performances.

\begin{figure}
    \centering
    \includegraphics[scale=.50]{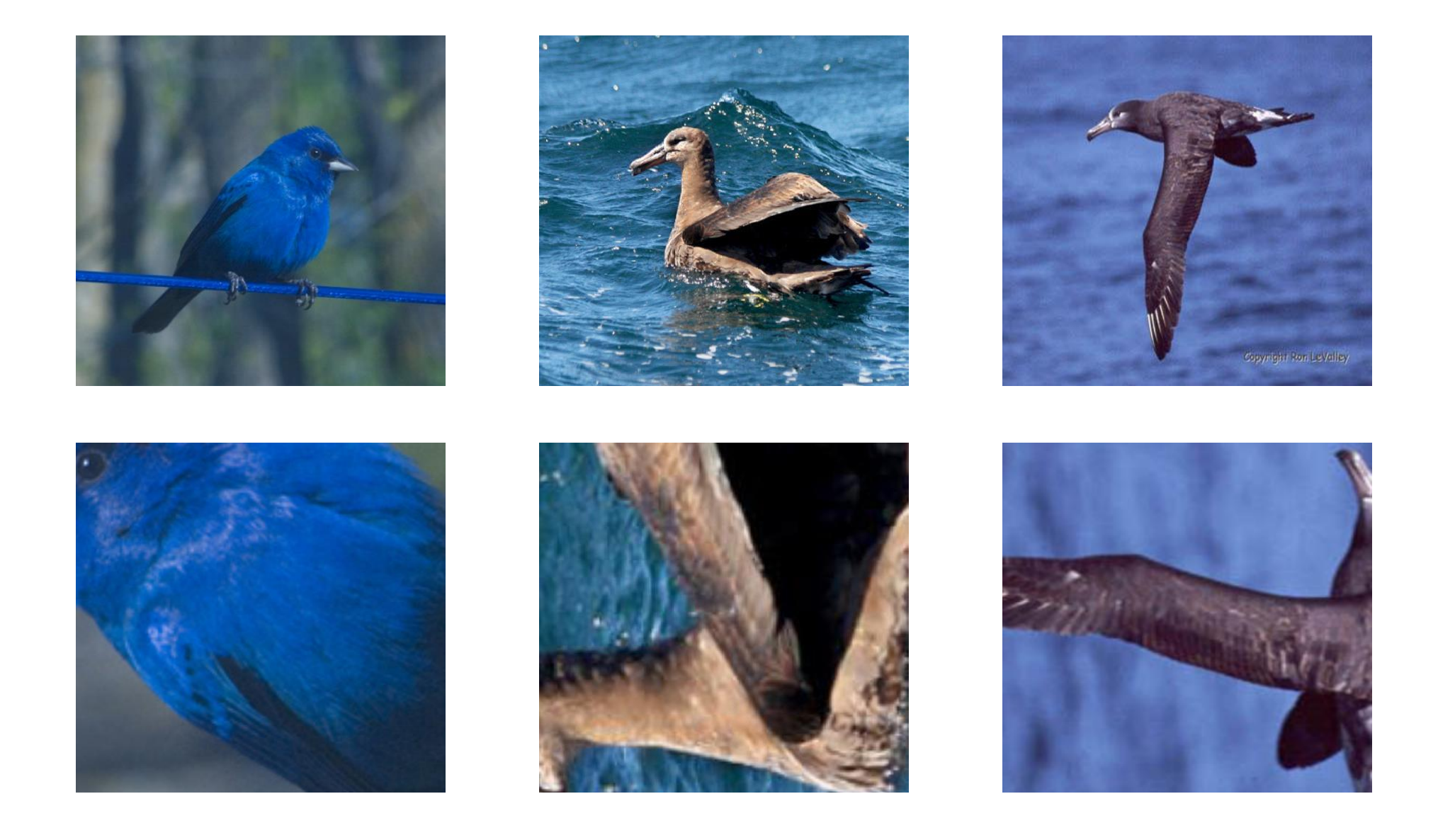}\vspace{-0.02in}
    \caption{Examples of original image and distorted image after projection transformation. The first row is original images, the second row is examples of distorted image after projection transformation}
    \label{fig:distored_img}
\end{figure}

Second, only few regular transformations have been adopted in the data augmentation in literature, such as translations, cropping and flips. This is not surprising -- many transformations would introduce sever distortions that would remarkably shear and crop images.
For example, Figure~\ref{fig:distored_img} illustrates some examples of such distorted images. These distorted images could be detrimental if they were naively used to train a model as they would introduce unrealistic features.

To address the above challenges, we propose a novel transformation-augmented regularization mechanism called FLAT (Few-shot Learning via Autoencoding Transformations) for the few-shot learning task. Specifically, we propose to use the idea of autoencoding transformations as a regularizer 
to learn better representations generalizable from the base to novel categories for few-shot learning tasks.  A family of transformations can be applied to regularize the representation learning of image features on base categories. 

Contrary to the naive data augmentation, the proposed transformation-based regularization will not directly train the representation network with the {\em supervised} labels of base-categories and the transformed examples. Instead, the randomly sampled transformations will be used to train the model in an {\em unsupervised} fashion, in hope to expand the feature space of base categories so that it could cover that of novel categories through transformation-augmented features without being limited to the base categories.
Thus, the unsupervised regularization on transformed images will enhance the model's generalizability by exploring the cross-category variations in the feature space of the novel categories \footnote{It does not mean an image would transform from one category to another. Instead, we expect it could induce and explore variations at {\em feature} level of transformations.}.

On the other hand, it also avoids the problem with distorted images since they are not directly used to train the representation network. Instead, the transformation-augmented images are used to inspect the possible changes of intrinsic visual structures under these transformations so that the learned representation can well encode and generalize with them. We will show its superior performances to the existing few-shot learning approaches, which sets new state-of-the-art results in literature.

The remainder of this paper is organized as follows. First, we will elaborate on the formulation in Section~\ref{sec:for}. Then we will review the related works in Section~\ref{sec:related}, followed by the experiments in Section~\ref{sec:exp}. Finally, we will conclude the paper in Section~\ref{sec:concl}.

\section{The Proposed Approach}\label{sec:for}

In this section, we will first discuss a simple yet effective baseline by pretraining over base categories
and fine-tuning over novel categories. Based on this baseline, a regularization mechanism is then presented to improve the generalizability of the model across different categories by exploring the feature-level variations induced by a distribution of transformations.

In a few-shot learning task, a set of labeled training examples are given on some base categories first.  Then the goal is to predict on novel categories where only few $K$-shot examples are labeled. 
\subsection{A Simple yet Effective Baseline}
A simple baseline few-shot learning approach is to pre-train a model in  a fully supervised fashion on the training set of base categories, and a classifier is retrained on top of the learned representation (usually the features from the beheaded neural network by removing the last output layer) by the pre-trained model.

Recent methods have shown such a simple algorithm can be very effective by initializing the last layer of the classifier of novel categories with the mean feature vector of $K$-shot examples, and fine-tuning it over the few examples. This baseline is called weight imprinting, which uses the feature mean as the prototype of each novel category. The approach has demonstrated competitive performances compared with the more advanced algorithms.

\subsection{Motivation: Autoencoding Transformation Regularization}
Unfortunately, the simple baseline does not explore the full ability of generalizing from training examples of base categories to predict on novel categories. One of the most significant challenges is the great potential of overfitting into the base categories, especially considering the fact that the base categories have overwhelming labeled examples compared with the few-shot examples on the target novel categories. Figure \ref{fig:overfitting} illustrates the overfitting problem when no suitable regularization mechanism is adopted.

Data augmentation plays a critical role in mitigating the overfitting problem by augmenting training examples with random transformations, such as translation, cropping and flipping. However, as aforementioned in Section~\ref{sec:intro}, it does not work for the few-shot learning task, since a naive data augmentation of training examples would either be limited to those of base categories, or have poor effect on novel categories with too few samples, let alone the adversarial distortions induced by more aggressively transformed images.

In particular, since the model is pre-trained on base categories, it is susceptible the pre-trained model only covers the semantic space of base categories. To avoid this problem, a novel regularization mechanism is needed to enhance the generalizability of the model from base to novel categories. This must be performed in an unsupervised fashion without naively using the labels of augmented examples to train the model. To this end, we propose the regularization of autoencoding transformations that seek to transform the {\em features} (rather than original images) beyond the semantic space of base categories to cover that of novel categories {\em without} knowing their labels.

\subsection{The Formulation}
Formally, consider an example $\mathbf x$ (e.g., from a base category) from data distribution $p(\mathbf x)$. We wish to learn an encoder $E_\theta$ with parameters $\theta$ as a {\em category generalizable} model to output the representation of $\mathbf x$. To train the encoder, the aforementioned baseline  uses the label $\mathbf y_{\rm base}$ of base categories as the target by learning a base classifier $C_{\psi_{\rm base}}$ with parameters $\psi_{\rm base}$ in a supervised fashion that minimizes the classification loss
\begin{equation}\label{eq:class}
\min_{\psi_{\rm base},\theta}\mathop \mathbb E\limits_{\mathbf x,\mathbf y_{\rm base}} \mathcal L (C_{\psi_{\rm base}}(E_\theta(\mathbf x)), \mathbf y_{\rm base})
\end{equation}
where the loss $\mathcal L$ is usually set to the cross-entropy loss between the predicted label and the ground truth.

However, the supervised training simply fits the classifier to the base categories whose underlying features may not well cover those of novel categories. Thus, we randomly sample a transformation $\mathbf t$ from a distribution $p(\mathbf t)$ of transformations, and apply it to a randomly drawn sample $\mathbf x \sim p(\mathbf x)$, in hope to generate augmented features $E_\theta (\mathbf t(\mathbf x))$ from the transformed image. However, such transformation-augmented features cannot be directly used to train the model as in Eq.~(\ref{eq:class}), as it still aims to fit with the base categories rather than exploring the variations within a larger set of categories involving novel classes. 

Thus, we resort to a transformation decoder, which instead aims to reconstruct the transformation. Unlike in the conventional autoencoders to reconstruct data, the transformation decoder $D_\phi$ seeks to recover the transformation $\mathbf t$ from the feature representations of original $\mathbf x$ and transformed images $\mathbf t(\mathbf x)$ by minimizing the deviation $\mathcal D$ between the estimated and original transformations
\begin{equation}\label{eq:trans}
\min_{\phi, \theta}\mathop \mathbb E\limits_{\mathbf x,\mathbf t} \mathcal D (\hat{\mathbf{t}} , \mathbf t),{\rm~~where~~}  \hat {\mathbf t} = D_\phi(E_\theta(\mathbf x),E_\theta(\mathbf t(\mathbf x)))
\end{equation}

Summing up Eq.~(\ref{eq:class}) and (\ref{eq:trans}), one could jointly optimize the representation encoder, transformation decoder and the base classifier, with a non-negative balancing coefficient $\lambda$,
\begin{equation}\label{eq:all}
\min_{\psi_{\rm base},\theta,\phi}\mathop \mathbb E\limits_{\mathbf x,\mathbf y_{\rm base}} \mathcal L (C_{\psi_{\rm base}}(E_\theta(\mathbf x)), \mathbf y_{\rm base}) + \lambda \mathop \mathbb E\limits_{\mathbf x,\mathbf t} \mathcal D (\hat{\mathbf{t}} , \mathbf t)
\end{equation}
where the regularization of reconstructing the transformation instructs the encoder to inspect the potential changes of feature representations without restricting them to just fit into the base categories as in Eq. (\ref{eq:class}). Thus, it encourages the generalization of transformed images beyond the base categories by exploring potential variations of features induced by a distribution of transformations from a rich family.

Once $E_\theta$ is learned, a novel classifier $C_{\psi_{\rm novel}}$ can be trained for predicting over the novel categories by minimizing
$$
\min_{\psi_{\rm novel}}\mathop \mathbb E\limits_{\mathbf x,\mathbf y_{\rm novel}} \mathcal L (C_{\psi_{\rm novel}}(E_\theta(\mathbf x)), \mathbf y_{\rm novel})
$$

The weights of the novel classifier can be imprinted by initializing to the mean of feature vectors over $K$-shot examples, which usually results in a classifier with good convergence.

\subsection{Discussions}

\begin{figure}
    \centering
    \includegraphics[width=0.9\textwidth]{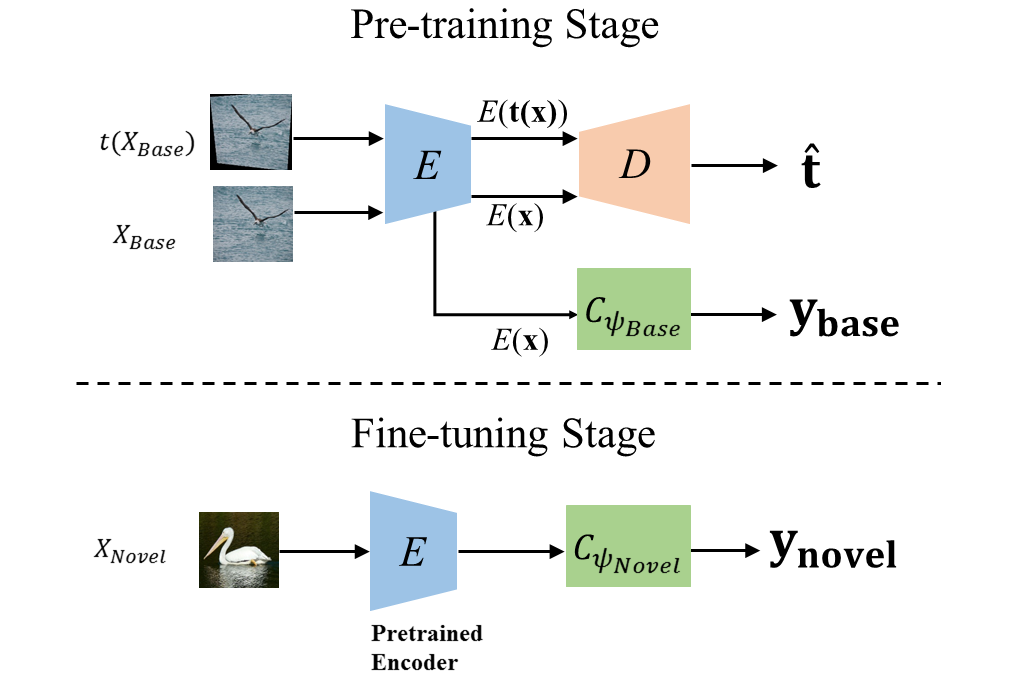}\vspace{-0.08in}
    \caption{Overview of two stages for training FLAT. The model pretrain on the base categories and fine-tune over the novel categories. In the pre-training stage, the classification error and transformation decoding loss are jointly minimized.}\vspace{-0.08in}
    \label{fig:overview}
\end{figure}

Figure~\ref{fig:overview} summarizes the two-stage training of the FLAT method -- the pre-training of the model on the  base categories, and the fine-tuning of it over the novel categories. In the pre-training stage, we jointly minimize the classification error and  transformation decoding loss with the data from base classes. In fine-tuning stage, we retrain a new classifier $\psi_{\rm novel}$ with the given labeled examples from novel classes. Like the baseline method \cite{qi2018low}, the weights of new classifier can be initialized to the mean of feature vectors over $K$-shot examples of novel classes.

\section{Related Works}\label{sec:related}
Few-Shot Learning (FSL) has been studied in many different ways. In particular, among them are many meta-learning based algorithms. One large category of meta-learning algorithms focus on learning the metrics to determine the semantic similarity between two samples so that a test sample can be classified based on the similarity with the few-shot examples \cite{sung2018learning,vinyals2016matching,snell2017prototypical}. The other category of meta-learning algorithms instead seek to learn the meta-knowledge about how to initialize \cite{finn2017model,jamal2018task} and optimize \cite{Ravi2017} the model. Both categories of meta-learning algorithms are trained by simulating the few-shot learning tasks over multiple episodes with the samples from base categories.

In contrast, weight imprinting and generators are presented to initialize and predict the weights of the last fully-connected layer for novel categories. For example, weight imprinting \cite{qi2018low} directly uses the mean feature vectors of few-shot examples to imprint the weights of novel categories. Alternatively, the weights can also be predicted as a function of the feature representations using their means \cite{qiao2018few} or an attention mechanism \cite{gidaris2018dynamic}.

Besides the few-shot learning algorithms, the proposed approach is also related with the autoencoding transformations \cite{zhang2019aet} recently proposed for unsupervised learning tasks.  It was used as an unsupervised algorithm to learn the representations of examples without predicting {\em any} labels. This differs from the proposed FLAT model in which the autoencoding transformation is used as regularizer to enhance the model's cross-category generalizability along with the minimization of the classification cost over base categories. It is the first time to show such a mechanism can play a significant role in regularizing the model training and delivering impressive classification performances across different categories for the few-shot learning tasks.

Finally, it is worth noting the difference from the conventional data augmentation methods applying some transformations to augment data \cite{krizhevsky2012imagenet}. The proposed FLAT does not use the augmented samples in a naive way to supervise the model training with their labels. Such a naive data augmentation would merely increase the risk of overfitting to base categories with their already overwhelming samples, or would not play any significant role if it was applied to novel categories with extremely few-shot examples. In contrast, without sticking to any labels of specific categories, the proposed FLAT would explore the variations induced by various transformations at the feature level to enhance its generalizability across different categories.

\section{Experiments}\label{sec:exp}
In this section, we evaluate the proposed FLAT model against several state-of-the-art methods on the CUB-200-2011 and miniImageNet datasets.
\subsection{Experiments on CUB-200-2011}\label{sec:exp_cub}
\subsubsection{Dataset configuration} The CUB-200-2011 dataset was originally proposed in \cite{cub2011} and contains 200 fine-grained categories of birds with 11788 images (about 30 training images per class on the average). Following train/test split method in \cite{qi2018low}, we use the standard train/test split provided by the dataset, and treat the first 100 classes as the base classes and the remaining 100 classes as the novel classes. Namely, we have 100 novel classes.

We strictly follow the evaluation setup in \cite{qi2018low} for a fair comparison, and all training examples sampled from novel classes are the same as used in \cite{qi2018low}. In the experiment, we set $K$ to $\{1, 2, 5, 10, 20\}$.

\subsubsection{Experimental settings} We conduct experiments in three settings on the CUB-200-2011 dataset in the same way as in \cite{qi2018low} for a fair comparison, including
\begin{itemize} \setlength\itemsep{0cm}
    \item All classes: the last layer of the base classifier $C_{\psi_{\rm base}}$ is expanded by imprinting and fine-tuning the weights of novel classes, leading to a 200-way classifier. We evaluate the model in a 200-way classification problem with all the test examples from both the base and novel classes.
    \item Novel classes: same as in the all classes setting, except we evaluate the model in a 200-way classification problem with the test examples from {\em only} the novel classes.
    \item Transfer learning: the last layer of the base classifier are replaced by imprinting and fine-tuning the weights of novel classes, leading to 100-way classifier. We evaluate the model in a 100-way classification problem on the testing examples from {\em only} the novel classes
\end{itemize}

\subsubsection{Compared methods} 
We compare against several state-of-the-art methods for few-shot learning, including Feature Generator~\cite{hariharan2017low} and Matching Networks~\cite{vinyals2016matching}. We follow \cite{qi2018low} to report the results of their proposed Imprinting+FT, as well as the baseline Rand+FT. 
Please refer to \cite{qi2018low} for more details of these methods.

\subsubsection{Architecture and Implementation Details}

\begin{figure*}
    \centering
    \includegraphics[scale=.60]{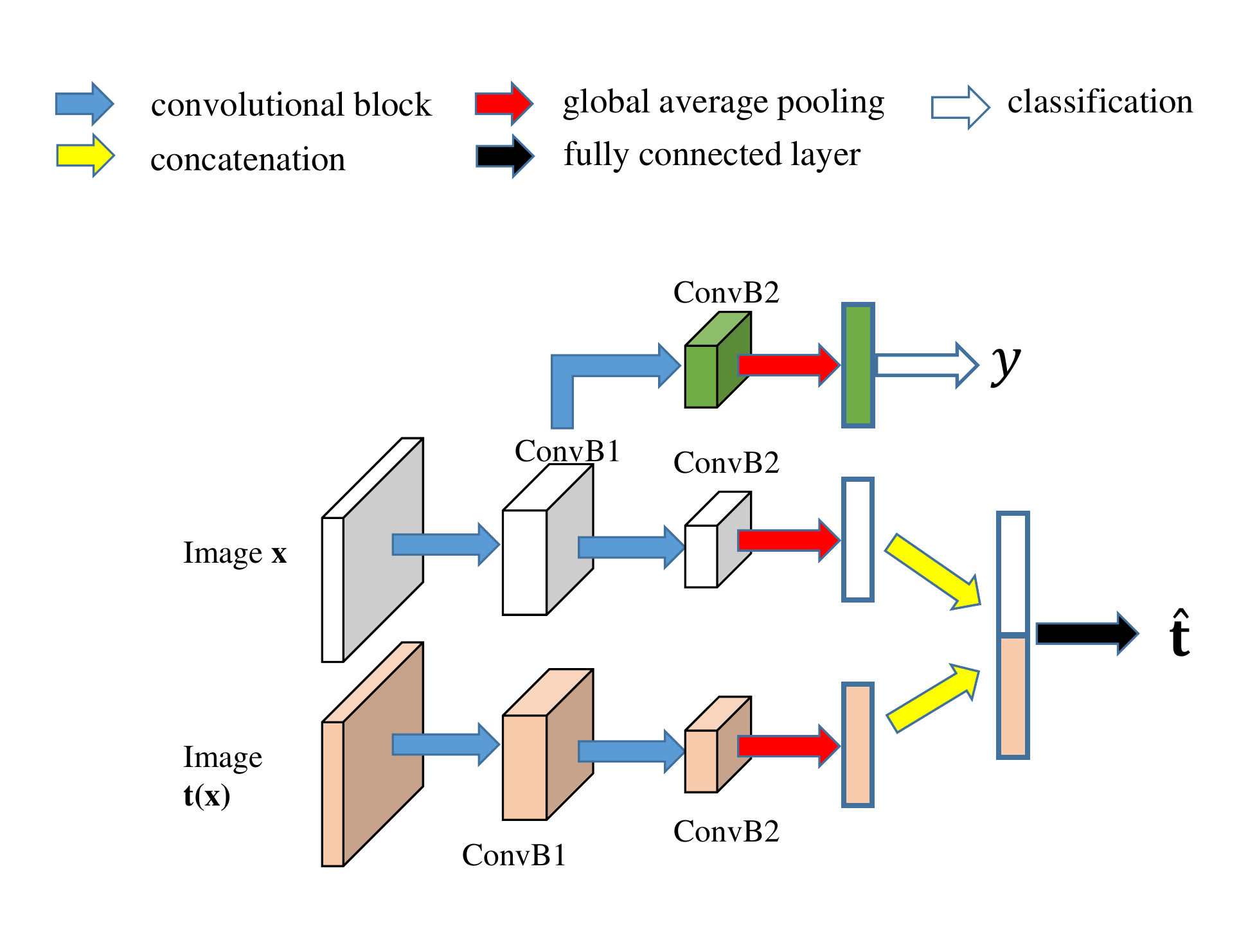}\vspace{-0.02in}
    \caption{An illustration of the network architectures for training FLAT with InceptionV1 backboneon CUB-200-2011}
    \label{fig:incep_arch}
\end{figure*}

To make a fair comparison with \cite{qi2018low}, we use InceptionV1~\cite{InceptionV1} as the network backbone. As illustrated in Figure \ref{fig:incep_arch}, the network consists of two branches with shared weights, each taking the original and the transformed images as input, respectively. The output features of the last block of two branches are averaged pooled and concatenated to form a 512-d feature vector. Then an output layer follows to predict the parameters of an input transformation.  In addition, the output features of the third last inception module in original input branch are connected to two more inception modules and a softmax classifier with one Fully-Connected(FC) layer to predict the label of an input image. No bias term is used in this layer. The overall training loss is the sum of the classification loss and the decoding transformation loss weighted by $4.0$ when pretraining on base classes through experiments.


During pre-training, all the convolutional layers are initialized from parameters pre-trained on the ImageNet dataset \cite{ILSVRC15} by convention.  The learning rate is set to 0.001 for the pre-trained layers with a $10\times$ multiplier used for the randomly initialized layers.  We apply an exponential decay at a rate of $0.1$ every 30 epochs, and train the model for a total of $3,00$ epochs with a batch size of 128 original images and their transformed counterparts. The SGD optimizer is adopted with a momentum of $0.9$ and a weight decay of $0.0001$. The projective transformation has shown competitive results in previous work on training transformation equivariant representations \cite{zhang2019aet}, and thus it is adopted in FLAT as well to minimize the transformation decoding loss. During fine-tuning, we set the learning rate to $0.0004$ and fine-tune the whole network for 140 epochs. 

Random cropping and horizontal flipping are also employed for data augmentation.

\subsubsection{Results on CUB-200-2011}

\begin{figure}
    \centering
    \begin{subfigure}{.49\textwidth}
    \centering
    \includegraphics[width=.95\linewidth]{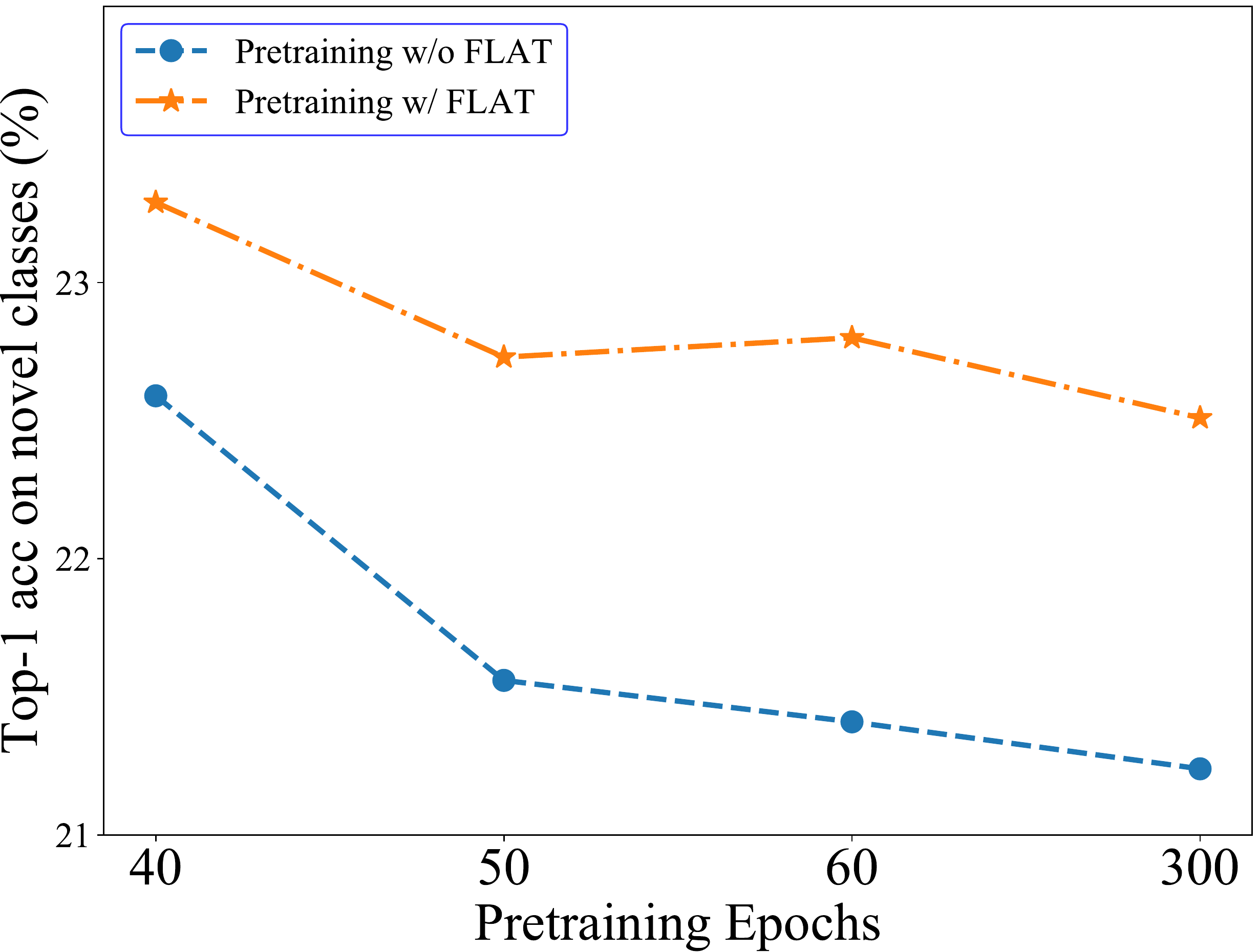}
    \caption{K=1}
    \end{subfigure}
    \hspace{0.01in}
    \begin{subfigure}{.49\textwidth}
    \centering
    \includegraphics[width=.95\linewidth]{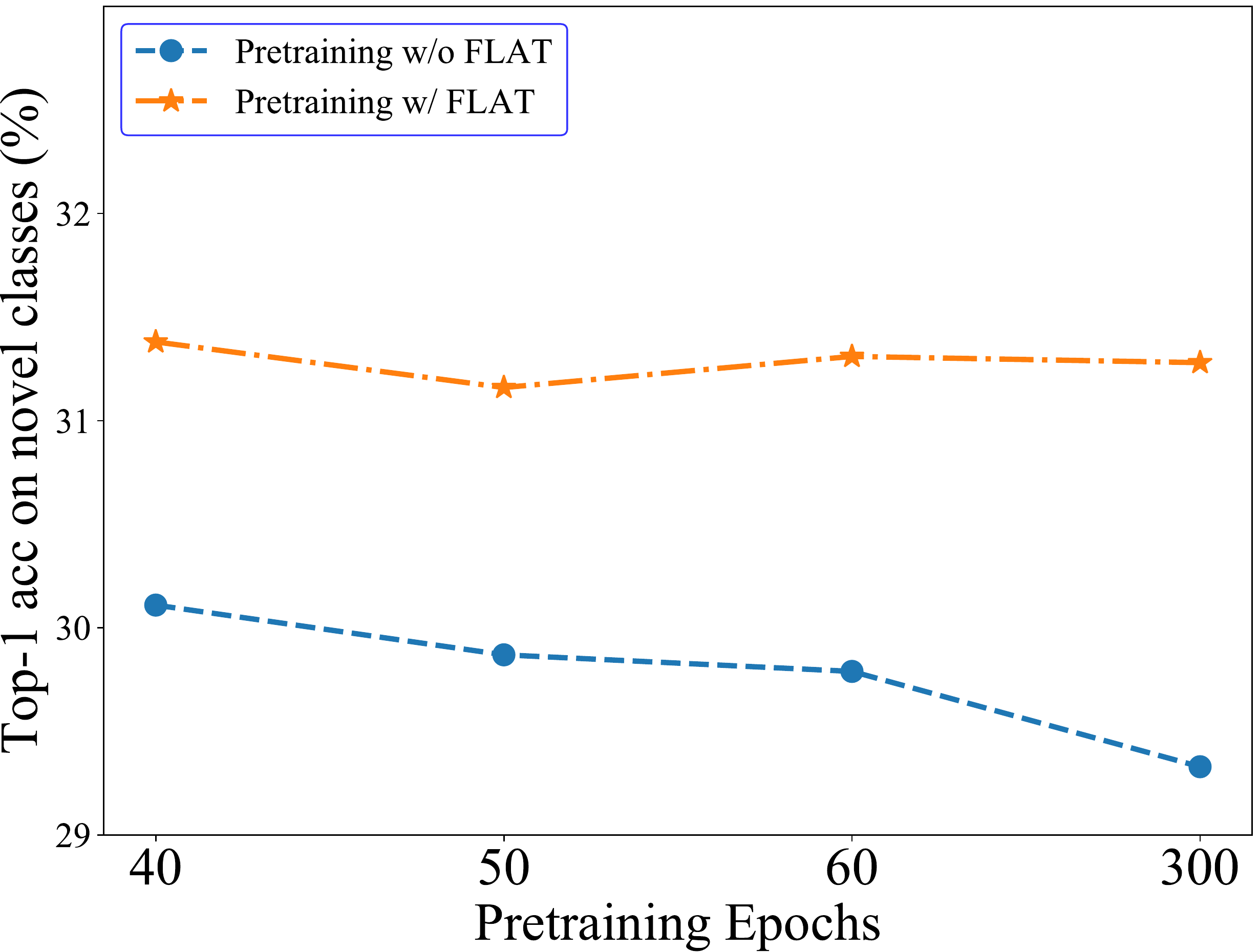}
    \caption{K=2}
    \end{subfigure}

\vspace{-0.08in}
\caption{Top-1 accuracy(in \%) by varying the pretraining epochs w/ and w/o FLAT. Pretraining with FLAT outperforms pretraining without FLAT, and  mitigates the overfitting problem significantly.}
\label{fig:overfitting}
\vspace{0.05in}
\end{figure}

In Figure \ref{fig:overfitting}, we compare the imprinting accuracy along with the pretraining epochs in novel class setting. Results show that pretraining  with  FLAT  outperforms  pretraining without FLAT, and mitigates the overfitting problem significantly. Tables \ref{tab:cub_all}, \ref{tab:cub_novel} and \ref{tab:cub_transfer} summarize the top-1 accuracy of the proposed methods and the compared methods in all the three experiment settings. In all settings, the proposed method consistently outperform all the variants of fine-tuning on novel classes from imprinted weights (Imprinting+FT) and random weights (Rand+FT) \cite{qi2018low}. This demonstrates the effectiveness of the proposed regularization mechanism. We also notice that applying projective transformation through data augmentation (Proj-transform+FT) leads to a poor performance. The poor performance reveals applying transformation through data augmentation does not work in this FSL scenario. As discussed in Section \ref{sec:intro}, many transformations may introduce distortions into images, these distorted examples can be even harmful as they would introduce unreal features if they were directly used to train the model.

\begin{table*}[t!]\fontsize{10}{10}\selectfont
    \caption{Top-1 accuracy (in \%) for the {\bf all classes} setting on CUB-200-2011 using 1--20 examples. Best results are shown in bold.}\label{tab:cub_all}
    \centering
        \setlength{\extrarowheight}{4pt}
    \begin{tabular}{clrrrrrr}
\hline
    $K=$ & 1 & 2 & 5 & 10 & 20\\
\hline
    Rand+FT \cite{qi2018low} & 39.26 &43.36 &53.69 &63.17 &68.75  \\
    Imprinting+FT \cite{qi2018low} & 45.81 &50.41 &59.15 &64.65 &68.73\\
    Proj-transform+FT &43.74 &48.47 &57.20 &63.47 &67.84 \\
    FLAT(ours) & {\bf47.31}& {\bf53.18} &{\bf62.03}& {\bf68.43} &{\bf72.26}\\
\hline
    Generator + Classifier \cite{hariharan2017low} &45.42 &46.56 &47.79 &47.88 &48.22\\
    Matching Networks \cite{vinyals2016matching} &41.71 &43.15 &44.46 &45.65 &48.63\\
\hline
\bottomrule
\vspace{-0.001in}
\end{tabular}

\end{table*}

\begin{table*}[t!]\fontsize{10}{10}\selectfont
\caption{Top-1 accuracy (in \%) for the {\bf novel classes} setting on CUB-200-2011 using 1--20 examples. Best results are shown in bold.}\label{tab:cub_novel}
    \centering
        \setlength{\extrarowheight}{4pt}

    \begin{tabular}{clrrrrrr}
\hline
    $K=$ & 1 & 2 & 5 & 10 & 20\\
\hline
    Rand+FT \cite{qi2018low} & 5.25 &13.41 &34.95 &54.33 &65.60\\
    Imprinting+FT \cite{qi2018low} & 18.67 &30.17 &46.08 &59.39 &68.77  \\
    Proj-transform+FT &18.51 &28.75 &45.54 &57.05 &64.88 \\
    FLAT(ours) & {\bf19.52}& {\bf31.88} &{\bf49.35}& {\bf62.01} &{\bf69.59}\\
\hline
    Generator + Classifier \cite{hariharan2017low} &18.56 &19.07 &20.00 &20.27 &20.88\\
    Matching Networks \cite{vinyals2016matching} &13.45 &14.75 &16.65 &18.81 &25.77\\
\hline
\bottomrule
\end{tabular}
\end{table*}

\begin{table*}[t!]\fontsize{10}{10}\selectfont
\caption{Top-1 accuracy (in \%) for the {\bf transfer learning} setting on CUB-200-2011 using 1--20 examples. Best results are shown in bold.}\label{tab:cub_transfer}
    \centering
        \setlength{\extrarowheight}{4pt}

    \begin{tabular}{clrrrrrr}
\hline
    $K=$ & 1 & 2 & 5 & 10 & 20\\
\hline
    Rand+FT \cite{qi2018low} & 15.90 &28.84 &46.21 &61.37 &71.57\\
    Imprinting+FT \cite{qi2018low} & 26.59 &34.33 &49.39 &61.65 &70.07  \\
    Proj-transform+FT &26.38 &32.84 &48.52 &57.05 &67.00 \\
    FLAT(ours) & {\bf28.19}& {\bf35.90} &{\bf51.16}& {\bf63.11} &{\bf71.67}\\
\hline
\bottomrule
\end{tabular}
\end{table*}

\subsection{Experiments on miniImageNet}\label{sec:exp_miniImageNet}

\begin{table*}\fontsize{11}{11}\selectfont
\caption{Top-1 accuracy (in \%) on miniImageNet with 95\% confidence interval. Best results are shown in bold.}\label{tab:miniimagenet}
    \centering\setlength{\extrarowheight}{4pt}
    \begin{tabular}{lrr}
\hline
    $K=$ & 1 & 5\\
\hline
    Rand+FT\cite{Ravi2017} & 28.86$\pm$0.54 & 49.79$\pm$0.79\\
    Nearest Neighbor\cite{Ravi2017} & 41.08$\pm$0.70 & 51.04$\pm$0.65\\
    \hline
    MAML~\cite{finn2017model} &  48.70$\pm$1.84 & 63.11$\pm$0.92\\
    Matching Network~\cite{vinyals2016matching} & 43.56$\pm$0.84 & 55.31$\pm$0.73\\
    Meta-Learner LSTM~\cite{Ravi2017} & 43.44$\pm$0.77 & 60.60$\pm$0.71\\
    Prototypical Networks~\cite{snell2017prototypical} & 49.42$\pm$0.78 & 68.20$\pm$0.66\\
    CovaMNet~\cite{LiAAAI2019} & 51.19$\pm$0.76 & 67.65$\pm$0.63\\
    DN4~\cite{LiCVPR2019} & 51.24$\pm$0.74 & 72.02$\pm$0.64\\
    \hline
    Activation-Simple~\cite{qiao2018few} &  54.53$\pm$0.40 & 67.87$\pm$0.20\\
    Activation-WRN~\cite{qiao2018few} & {59.60$\pm$0.41} & 73.74$\pm$0.19\\
    Imprinting+FT~\cite{qi2018low} & 57.40$\pm$0.81 & 75.15$\pm$0.63\\
    \hline
    FLAT (ours) &{\bf59.88$\pm$0.83} &{\bf77.14$\pm$0.59}\\
\hline
\end{tabular}
\end{table*}

\subsubsection{Dataset Configuration}
The miniImageNet dataset, originally proposed in \cite{Vinyals2015}, has been widely used for evaluating few-shot learning algorithms. It consists of $60,000$ color images from $100$ classes with $600$ examples per class. It is a simplified subset of ILSVRC 2015~\cite{ILSVRC15}. We use the train/val/test split introduced in \cite{Ravi2017} and follow its evaluation protocol. In particular, we use the 20 test classes as the novel classes and the rest 80 classes as the base classes, and the evaluation is only performed on novel classes. We repeat the experiments 600 times and report the mean accuracy with 95\% confidence interval. In each experiment, we randomly sample five classes from novel classes, and for each class, we randomly select one or five labeled images as our support set and $15$ examples for the query set.

\subsubsection{Compared methods}
The miniImageNet dataset has been widely used for evaluating few-shot learning algorithms, and is a good benchmark to compare with several sate-of-the-art methods such as Matching Networks~\cite{vinyals2016matching}, MAML~\cite{finn2017model}, Prototypical Networks~\cite{snell2017prototypical}, and some baseline methods such as {Rand+FT}, {Nearest Neighbor}. We also report the results of CovaMNet~\cite{LiAAAI2019} and DN4~\cite{LiCVPR2019}, as well as two variants of the methods proposed in \cite{qiao2018few}, i.e.,~Activation-Simple and Activation-WRN. In addition, we also evaluate Imprinting+FT~\cite{qi2018low} on the miniImageNet dataset.

\subsubsection{Architecture and Implementation details}
To compare with the baselines in \cite{qiao2018few}, we adopt the wide residual network, {i.e.,~WRN-28-10}~\cite{WRN} as the backbone. To speed up training process, all the convolutional layers are initialized with parameters pre-trained on the base classes of miniImageNet. We also made a modification to accept input of a different size. The input image is cropped to $80\times80$. Like in the architecture introduced in Section \ref{sec:exp_cub},
the backbone network is modified by adding the transformation decoder upon the last block, and

the output features of the last second WRN basic layer in original input branches are average pooled and connected to a softmax classifier with one Fully-Connected(FC) layer to predict labels of input images.

The similar learning process (e.g., learning rate scheduling, network optimizer, weight decay, and loss trade-off coefficient) used in the experiments on the CUB dataset is adopted here to train the base model for a total of 90 epochs.

For a fair comparison, we follow the standard evaluation protocol for this dataset by considering the transfer learning setting only~\cite{Ravi2017,snell2017prototypical,vinyals2016matching}, and using the few-shot examples from novel classes for fine-tuning the model for a total of 15 epochs with a batch size of 32, and a learning rate of $0.002$. 

\subsubsection{Results on miniImageNet}
The results are summarized in Table \ref{tab:miniimagenet}, where the proposed  FLAT outperforms the other methods. The competitive performance demonstrates the advantage of the FLAT by introducing the regularization of autoencoding transformations on learning image representations. It successfully learns a generalizable representation for novel classes with few-shot examples. The FLAT as a base representation model can be integrated into the other few-shot learning methods to enhance their generalizability on novel classes. However, we merely adopt basic methods to show the critical role of self-supervised autoencoding regularizer in few-shot learning.

\subsection{Experiments on Cross-Dataset Scenario}
\begin{table*}\fontsize{11}{11}\selectfont
\caption{5-shot accuracy under the cross-dataset scenario with a ResNet-18 backbone. Best results are shown in bold.}\label{tab:cross_dataset}
\centering\setlength{\extrarowheight}{3pt}
\begin{tabular}{lr}\\\hline
&mini-ImageNet$\rightarrow$ CUB \\
\hline
Baseline\cite{chen2019closer} & 65.57$\pm$0.70\\  
Baseline++\cite{chen2019closer} & 62.04$\pm$0.76\\ 
\hline
    ProtoNet\cite{snell2017prototypical} &62.02$\pm$0.70\\
    MatchingNet\cite{vinyals2016matching} &53.07$\pm$0.74\\
    MAML\cite{finn2017model} &51.34$\pm$0.72\\
    RelationNet\cite{sung2018learning} &57.71$\pm$0.73\\
\hline
    FLAT (ours) &{\bf67.69$\pm$0.68}\\
\hline
\end{tabular}
\end{table*}

Cross-Dataset experiment for few-shot learning has been evaluated in \cite{chen2019closer}. It is a  practical yet more challenging scenario since the images and their classes could be quite different from each other across datasets.
In addition, compared with standard evaluation methods, cross-dataset experiment can better reveal the generalizability of the base model across datasets.
\subsubsection{Dataset Configuration} We use mini-ImageNet to form our base class, and evaluate on 50 novel class from CUB. Class split methods are same as \cite{chen2019closer} for a fair comparison. 
\subsubsection{Compared methods}  We compare with  the Baseline \cite{chen2019closer}, Baseline++ \cite{chen2019closer}, as well as Matching Networks\cite{vinyals2016matching}, MAML\cite{finn2017model},RelationNet\cite{sung2018learning}, ProtoNet\cite{snell2017prototypical} in literature.

\subsubsection{Architecture and Implementation details}
Following \cite{chen2019closer}, we use the ResNet-18 as the backbone. We follow exactly the same training and testing process as \cite{chen2019closer}.  In fine-tuning stage on novel classes, we retrain a classifier from scratch on top of the learned representation, instead of retraining from imprinted weights \cite{qi2018low}. The networks consists of two branches with shared weights, each taking the original and the transformed images as input, respectively. The output features of the last block of two branches are averaged pooled and concatenated to form a 512-d feature vector to predict the parameters of input transformation. The output features of the last second resnet layer in original input branch are average pooled and connected to a softmax classifier with one Fully-Connected(FC) layer to predict labels of input images.

During pre-training, we use the similar process as in the experiments on miniImageNet to train the model classifier for a total number of $85$ epochs.

During fine-tuning, the same process for the miniImageNet experiments is also adopted to fine-tune the classifier for a total of $100$ epochs, and we follow \cite{chen2019closer} to update $\psi_{\rm novel}$ on novel CUB classes from scratch. 

\subsubsection{Results on Cross-Dataset Scenario}
The results are summarized in Table \ref{tab:cross_dataset}. the proposed methods FLAT outperforms the other compared methods. The FLAT gains $2.1\%$ relative improvement over the best performing baseline method \cite{chen2019closer}. It shows that the proposed method improves the generalizability of the model across different dataset, and is less affected by domain shift across datasets.

\section{Conclusions}\label{sec:concl}

This paper presents a novel FLAT (Few-shot Learning via Autoencoding Transformations) regularization mechanism to explore the variations in feature representations induced by a family of transformations. It aims to enhance the generalizability of the model from the base to the novel categories, which encourages the representation encoder to inspect the potential changes of transformation-augmented features without being limited to base categories. Finally, the superior performances in the experiments demonstrate the learned model could well generalize to cover the semantic space of novel categories by exploring the variations of transformed features.

%


\small
\bibliographystyle{plain}
\bibliography{fsl}

%
%
%

\end{document}